\newcommand{\name}{\texttt{DESYR}}
\documentclass[sigconf]{acmart}
\usepackage{amsmath}
\usepackage{multirow}
\usepackage{arydshln}
\usepackage{soul,xcolor}
\usepackage{enumitem}

\usepackage[ruled]{algorithm2e} 

\AtBeginDocument{%
  \providecommand\BibTeX{{%
    \normalfont B\kern-0.5em{\scshape i\kern-0.25em b}\kern-0.8em\TeX}}}

\copyrightyear{2021}
\acmYear{2021}
\setcopyright{acmcopyright}
\acmConference[CIKM '21]{Proceedings of the 30th ACM International Conference on Information and Knowledge Management}{November 1--5, 2021}{Virtual Event, QLD, Australia}
\acmBooktitle{Proceedings of the 30th ACM International Conference on Information and Knowledge Management (CIKM '21), November 1--5, 2021, Virtual Event, QLD, Australia}
\acmPrice{15.00}
\acmDOI{10.1145/3459637.3482423}
\acmISBN{978-1-4503-8446-9/21/11}

\begin{document}

\title{DESYR: Definition and Syntactic Representation Based Claim Detection on the Web}

 
\author{Megha Sundriyal$^1$*, Parantak Singh$^2$*, Md Shad Akhtar$^1$\\ Shubhashis Sengupta$^3$, Tanmoy Chakraborty$^1$ }
\thanks{* First two authors have equal contributions. The work was done when Parantak was an intern at LCS2 Lab, IIIT-Delhi.}

\affiliation{
  \institution{
                $^1$\textit {IIIT-Delhi, India.}
                $^2$\textit{BITS Pilani, Goa, India.} 
                $^3$\textit{Accenture Labs Bangalore, India.}
                }
    \country{\textit{India}}
                }
              
\email{{meghas, shad.akhtar, tanmoy}@iiitd.ac.in,} 
\email{parantak.singh@gmail.com, shubhashis.sengupta@accenture.com}





\begin{abstract}
The formulation of a claim rests at the core of argument mining. To demarcate between a claim and a non-claim is arduous for both humans and machines, owing to latent linguistic variance between the two and the inadequacy of extensive definition-based formalization. Furthermore, the increase in the usage of online social media has resulted in an explosion of unsolicited information on the web presented as informal text. To account for the aforementioned, in this paper, we propose \name. It is a framework that intends on annulling the said issues for informal web-based text by leveraging a combination of hierarchical representation learning (dependency-inspired Poincaré embedding), definition-based alignment, and feature projection. We do away with fine-tuning compute-heavy language models in favor of fabricating a more domain-centric but lighter approach. Experimental results indicate that \name\ builds upon the state-of-the-art system across four benchmark claim datasets, most of which were constructed with informal texts. We see an increase of 3 claim-F1 points on the LESA-Twitter dataset, an increase of 1 claim-F1 point and 9 macro-F1 points on the Online Comments (OC) dataset, an increase of 24 claim-F1 points and 17 macro-F1 points on the Web Discourse (WD) dataset, and an increase of 8 claim-F1 points and 5 macro-F1 points on the Micro Texts (MT) dataset. We also perform an extensive analysis of the results. We make a 100-D pre-trained version of our Poincaré-variant along with the source code.
\end{abstract}




\keywords{Claim Detection, Poincaré Embedding, Feature projection, Informal texts, Linguistic Grounding, Twitter, Social media}

\maketitle
 {\fontsize{8pt}{8pt} \selectfont
\textbf{ACM Reference Format:}\\
Megha Sundriyal, Parantak Singh, Md Shad Akhtar, Shubhashis Sengupta, Tanmoy Chakraborty. 2021. DESYR: Definition and Syntactic Represen-tation Based Claim Detection on the Web. In {\it Proceedings of the 30th ACM International Conference on Information and Knowledge Management (CIKM'21), November 1-5, 2021, Virtual Event, QLD, Australia}. ACM, New York, NY, USA, 10 pages. https://doi.org/10.1145/3459637.3482423 }

\vspace{-1mm}
\section{Introduction}
The increasing vogue behind Online Social Media (OSM) has led to a colossal sway within the media-consuming audience. Participation over OSM has swiveled into another corresponding and associate phenomenon. We are incessantly exposed to news, media, opinions, and perspectives. Such a constant barrage of information can, in turn, increase the plausibility behind resources being mishandled or exploited. 

The 45$^{th}$ Presidential US elections witnessed the alarming impact of fake news where a substantial percentage of their citizens were influenced by a malicious website \cite{grave2018learning}. In their study, \citet{allcott2017social} revealed that every American clicked on at least one fake news article related to the presidential candidates during the elections. The recent cases of spreading fear and prompting false cures due to fake news spread have threatened countless lives during the COVID-19 global pandemic \cite{un2020during}. The effect of claims applies unprecedentedly and often leads to the loss of money and precious human life. Similarly, several such cases have been popping up within the global community. Several machine learning and natural language processing models have been proposed to handle fake news and the automated detection of erroneous claims. Within recent literature, fake news detection has been embellished into more of an umbrella term that encapsulates sub-tasks like stance detection, hostility detection, claim detection, etc.

Argument Mining (AM) has been significantly confronted by NLP for the past few decades.  Segmenting argumentative and non-argumentative texts, detecting claims, and parsing argument assemblies are some of the main concentrations within this field. Narrowing down from AM into the domain of claim detection, the task in layman terms is to detect sentences containing a claim. As an elementary intuition, claim detection is a pre-cursor to fact-checking, wherein segregation of claims aids in restricting the corpus that needs a fact-check.

\citet{toulmin2003uses} initiated the early works on argumentation in the 1950s; in his argument conjecture, he described a \textit{claim} as `{\em an assertion that needs to be proven}'. Formally, according to the Oxford Dictionaries , a claim is defined as: \textit{"to say that something is true although it has not been proven and other people may not believe it"} \footnote{\href{https://www.oxfordlearnersdictionaries.com/definition/english/claim_1?q=claim}{Claim - Oxford Definition}}. The concept of false claims is entrenching its roots in every domain: online social media platforms, news articles, political agendas, and so forth. The divergence boundary between a claim and a non-claim is very thin, making the conception of a claim very subjective and abstruse. As a consequence, the categorization of claims is taxing for both human annotators and state-of-the-art neural models alike. Herein, the difficulty exists given the disparity in perception and the lack of an existing formalization for claims. The more classical works in claim detection practiced syntactic composition, where they cashed in on the use of combinatorial distillation from context-free grammars, constituency parse-trees, and other linguistic renditions to design their models \cite{levy2014context, lippi2015context}. The more recent works shifted towards neural approaches, and leveraging large Language Models (LM) in an attempt at capturing dormant features through explicit linguistic encapsulation \cite{daxenberger2017essence, chakrabarty2019imho}. In our previous work \cite{gupta-etal-2021-lesa}, we attempted to encode both linguistic and contextual knowledge to study claims across different distributions. The literature across the encompassing of OSM-based informal texts and the fusion of structure and context for claim detection is still sporadic and needs to be driven. Table \ref{tab:examples} shows a few representative examples of claims and non-claims. 

\begin{table}[t]
    \centering
    \caption{Representative examples of claims and non-claims. Explicit expressions of claims are in italics.}
    \label{tab:examples}
    \begin{tabular}{p{20em}|c}
        \bf Text & \bf Label \\ 
        \hline
        
        \hline
        @realDonaldTrump A lot of people are saying \textit{cocaine cures COVID-19.} So. & Claim \\ \hline
        \textit{Injecting disinfectant might not cure \#coronavirus, but TheLastDance sure is saving lives during \#Lockdown.} & Claim \\ \hline
        maybe if i develop feelings for covid-19 it will leave & Non-claim \\ \hline
        Olympic events are rooted in old traditions. & Non-claim \\ \hline
    \end{tabular}
\end{table}

\subsubsection*{\bf Motivation:} Claims, as is very apparent, subsist across a variety of distributions, from essays, to Wikipedia articles, to OSM posts and comments, etc. However, with the voluminous explosion of data on social media, it is of paramount importance that we concentrate distinctively upon claims distributed through OSM \cite{baum2021covid, world2020immunizing}. The web has become the pivot for all things social and global, and a vast majority of this human expression comes in textual form, specifically short, informal texts (Twitter, Facebook, etc.). More often than not the \textit{conformity bias} \cite{whalen2015conformity} comes into play and pits users of strong wide-ranged opinions against each other, and can at times end up promoting something erroneous. For example, a tweet that reads \textit{`drinking an antiseptic cures COVID'} can be a cause for massive unrest and demands an immediate fact check. Visibly, with the increasing magnitude of data, automated furthering of texts that comprise a claim into a fact-verification pipeline should garner gravity.

Another qualm, as previously mentioned, is the issue of division between claims and non-claims. These, at times, tend to lie in a domain of high invariability. Additionally, the severe imbalance of existing datasets leads to data severance (for performance augmentation) within already small datasets. On top of this, existing systems have inclined towards large LMs and provided a profusion of promise. However, they are, without a doubt, computationally expensive. We address the following intricacies and propose a framework that pursues the objective of erecting discernible feature spaces for the individual classes while avoiding the use of LMs and fixating on a linguistic and definition centring approach.

\subsubsection*{\bf Proposed Methodology:} We propose \textbf{\name}, a \textbf{DE}finition and \textbf{SY}ntactic \textbf{R}epresentation based claim detection model that achieves competence over the better segregation of feature space for classification, and moreover, learns to leverage the guidelines for identifying claims and non-claims proposed in our previous work \cite{gupta-etal-2021-lesa} for furtherance of the feature constitution. It attains this by employing an intelligible unity of feature projection, attention-based alignment, and pre-transformer Deep Learning.
Also, in line with \citet{lippi2015context} who argued that a claim could be grounded in linguistics, we propose \textbf{DARÉ}, the \textbf{D}ependency-Poinc\textbf{ARÉ} embedding, which is a dependency-inspired variant of Poincaré embedding \cite{nickel2017poincare}. It helps assimilate enhanced representations of word vectors by capturing intrinsic hierarchies in dependency trees.

We evaluate \name\ across four web-based datasets which comprise short informal texts and observe up to par results throughout. We conduct a result analysis across several baselines, along with a general analysis for our predictions. The comparative investigation espouses the finer performance of \name\ compared against other state-of-the-art systems.

\subsubsection*{\bf Summary of the major contributions:}
\vspace{-0.1cm}
\begin{itemize}[leftmargin=*]
    \item \textbf{\name, a claim detection system tailored towards informal web-based texts}. The said system determines the existence of claims in online text by aligning the query to encoded definitions and projecting them into a purer space. 
    
    \item \textbf{Novel combination of gradient reversal layer and attentive orthogonal projection.} To the best of our knowledge, the usage of gradient-reversal layer to learn the label-invariant features and leveraging an attentive orthogonal projection for drawing the choicest feature representations in the domain of claim detection is the primary attempt.  
    
    \item \textbf{Comprehensive evaluation and state-of-the-art results.} We evaluate \name\ against the state-of-the-art systems and LM baselines across four benchmark datasets. The contrast suggests the superiority of our architecture, and our ablation study highlights the importance of each module.
    \item \textbf{DARÉ, our novel dependency-inspired Poincaré variant}. It shows promise for linguistically-grounded NLP tasks. We release a pre-trained 100-dimensional version of DARÉ.
\end{itemize}

\subsubsection*{\bf Reproducibility:} We release the 100-D pre-trained version of our Poincaré-variant (DARÉ) and source code of \name\ for further research at {\url{https://github.com/LCS2-IIITD/DESYR-CIKM-2021}}.

\begin{figure*}[!h]
	\centering
    \includegraphics[width=0.95\textwidth]{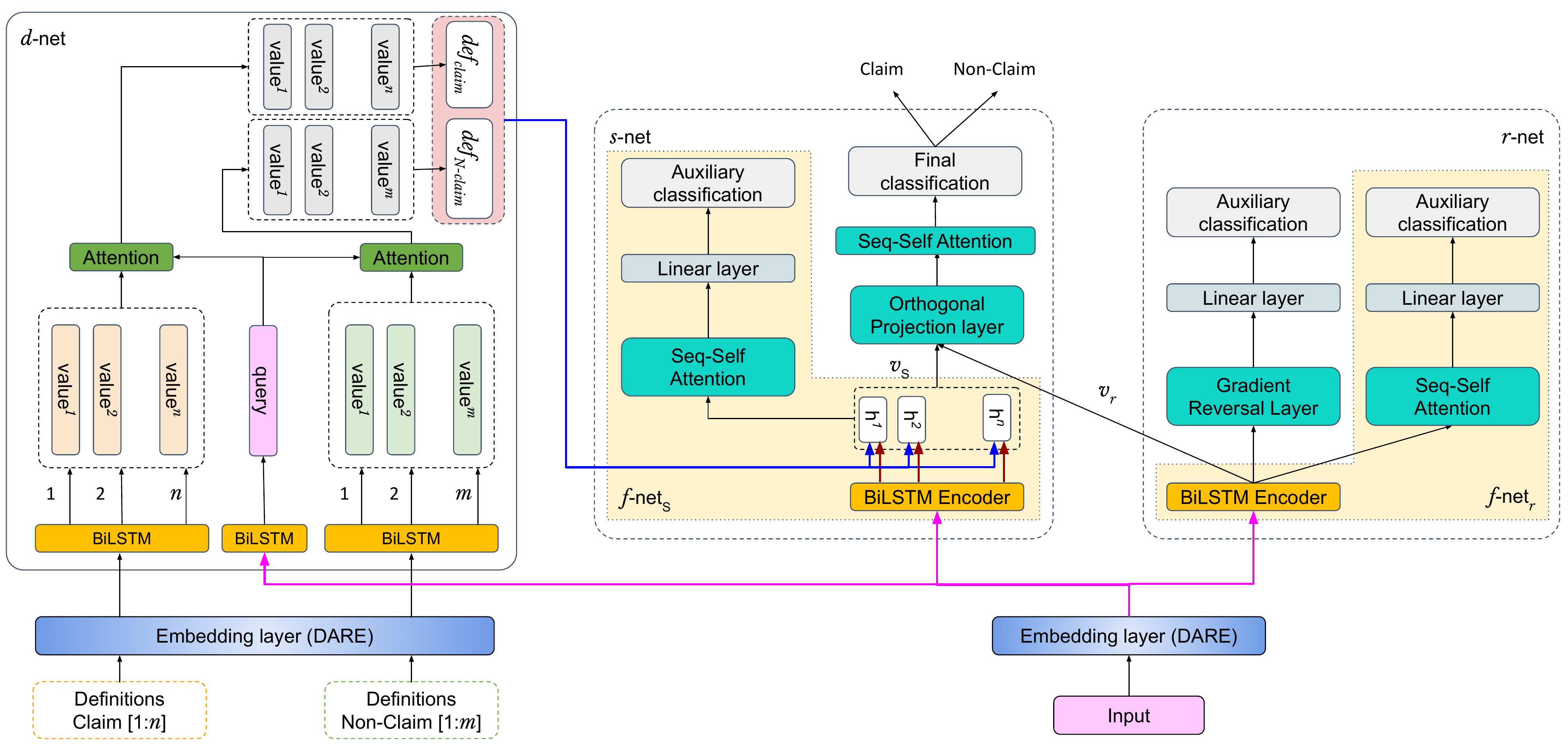}
    \caption{A schematic diagram of the \name\ framework for the claim detection. The spotlight network, \textit{s-net}, is the backbone of \name\ that incorporates the alignment of input text considering the claim and non-claim definitions. Moreover, it selectively extracts the relevant features through an attentive orthogonal projection of label-invariant features (from \textit{r-net}) and the aligned input representation (from {\em d-net}). The auxiliary classification layers help \name\ in learning sub-modules directly from the gradients' signals.}
    \label{fig:model}
\end{figure*}

\section{Related Work}
Claims are an indispensable component of AM, and similarly claim detection is equally as crucial for the fake news detection pipeline. With the expansion of OSM, misinformation proliferated through claims now undeniably poses a greater risk to consumers. With zero deterrence, the spread of misinformation can be rapid and harmful. Therefore, claim detection has recently gained traction as an NLP task which works a precursor to automated fact verification. 

In 2011, \citet{bender-etal-2011-annotating} proposed an annotation system and presented the Authority and Alignment in Wikipedia Discussions (AAWD) corpus\footnote{http://ssli.ee.washington.edu/projects/SCIL.html} which consisted of a collection of around 365 discussions curated from Wikipedia Talk Pages (WTP). Their work grabbed a lot of attention from researchers on claims and provided a basis for this challenging task. In the past decade, the study of claim detection procured some drag within the NLP research community with a principal attempt by \citet{rosenthal2012detecting}. They endeavoured on mining claims from discussion platforms and implemented a supervised approach based on sentiment and word-gram derivatives. Although their work was restricted to a classical machine learning approaches, it formed the basis for future works in this field.

\citet{levy2014context} put forward a context-dependent claim detection (CDCD) model. They described a `context-dependent claim' as \textit{"a general, concise statement that directly supports or contests the given topic"}. They exercised their approach only over Wikipedia scripts while maneuvering context-based and context-free feature sets for spotting claims. \citet{lippi2015context} propounded the context-independent claim detection (CICD) model that employed constituency parse trees to capture structural knowledge without an explicit encapsulation of context. The pitfall being that they only engineered this around a thoroughly superintended Wikipedia corpus.

\citet{levy2017unsupervised} proposed the first unsupervised approach for claim detection. According to the authors, a claim begins with the word \textit{`that'} and the main concept (MC) or a topic name then follows. This work, however, is restrictive to distributions which are accompanied by formal texts.  Most literature on claim focuses focuses on the domain specificity. As a result, there is a lack of generalization in existing systems. To confront this, \citet{daxenberger2017essence} performed a qualitative analysis across six datasets and argued that the anatomy of claims stands differently within different distributions. 

In the more recent times, the study of claims too, trended towards the utilization of transformers. \citet{chakrabarty2019imho} used over 5 million self-labeled Reddit comments that contained the abbreviations IMO (In My Opinion) or IMHO (In My Honest Opinion) to fine-tune their LM expecting to gravitate their distribution towards their conceptualization of a claim. They however, made no evident attempt at encapsulating the syntactic properties. \citet{gupta-etal-2021-lesa} leveraged both semantic and latent syntactic features through an amalgamation of linguistic encoders (part-of-speech and dependency based) and a contextual encoder (BERT). They additionally annotated a Twitter dataset and proposed thorough guidelines that were centred around annotating claims. \citet{cheema2021role} investigated the role of images on claims. They presented a novel framework on leveraging dual modalities for claim detection.

The CLEF-2020 shared task \cite{barron2020checkthat} witness several models that were specifically tweaked for claim detection. \citet{williams2020accenture} bagged the cake with the fine-tuned RoBERTa \cite{liu2019roberta} model that was accentuated by mean pooling and dropout. \citet{nikolov2020team} took the second place with their out-of-the-box RoBERTa vectors that were heightened by Twitter metadata. 

\name\ attempts at abrogating the drawbacks from previous works - we propose an architecture grounded in linguistics and expedited by context and definition-based alignment.


\section{Proposed Methodology: \name}
The more conventional renderings of claim detection are constructed with either contextual methodologies \cite{chakrabarty2019imho, daxenberger2017essence} or syntactic methodologies \cite{levy2017unsupervised, lippi2015context} at the centre. Recently, in our precedent work \cite{gupta-etal-2021-lesa}, we proposed the LESA architecture that leverages individually-furnished contextual and linguistic encoders combined into one for improved performance on claim detection. Moreover, in the helm of natural language processing, we observe increasing use of language models and fine-tuning in addition to other modules in the architecture -- a computationally heavy process with a significant number of learnable parameters. In LESA \cite{gupta-etal-2021-lesa}, we incorporated three modules including two transformer-based models for the claim detection, thus making the architecture heavy on compute. Additionally, within the literature, we observe an inferior performance on the minority class, despite efforts on imbalance eradication. Given our outlook on the assimilation of syntax and context and driven by an attempt to prune the aforementioned drawbacks, we now propose \name.

\name\ leverages representation learning by exercising a novel dependency-inspired variant of the Poincaré embedding \cite{nickel2017poincare}. Furthermore, we aim at amputating the class-invariant features, and obtaining superior class-representing features by incorporating the technique for the feature projection, highlighted by  \citet{qin2020feature}. \name's backbone comprises of two networks in parallel -- the regulation-net (\textit{r-net}) and the spotlight-net (\textit{s-net}). The regulation-net learns the class-invariant features through a gradient-reversal layer \cite{ganin2016domain}. On the other hand, the spotlight-net draws representations of the input which are devoid of the class-invariant features, thereby allowing for better distinction between our binary labels.
Both \textit{r-net} and \textit{s-net} incorporate a feature extractor module, called feature-net (\textit{f-net}), in their respective modules -- \textit{f-net}$_s$ and \textit{f-net}$_r$. 
In addition, we also incorporate a definition network (\textit{d-net}) that aims at aligning input texts to definitions of a claim and non-claim to further augment class variance. We leverage \textit{d-net} in our spotlight network module by enabling the feature network (\textit{f-net}$_s$) to learn the segregation between claim and non-claim definitions. We hypothesize that learning such alignment would be exploited by \name\ in successive layers for the claim detection. In particular, we fine-tune the hidden representation in \textit{s-net} to extract the essence of segregation between the claim and non-claim definition through {\em d-net}.

Moreover, \name\ is specifically calibrated towards short informal web-based texts, and we leverage claim and non-claim definitions proposed in \cite{gupta-etal-2021-lesa} to engineer our \textit{d-net}. Intricate details of the aforementioned modules are discussed in further sections. We present a high-level architecture for \name\ framework in Figure \ref{fig:model}. 

\subsection{Dependency-Poincaré Embedding (DARÉ)}
The Depedency-poincARÉ (DARÉ) embedding is \name's variant of the Poincaré embedding  \citet{nickel2017poincare}. Employing DARÉ, we attempt to capture latent linguistic properties in textual dependency. Below, we briefly introduce dependency parsing followed by a dossier on how the former was imbued into the Poincaré ball.

\begin{figure}[t]
	\centering
    \includegraphics[width=0.4\textwidth]{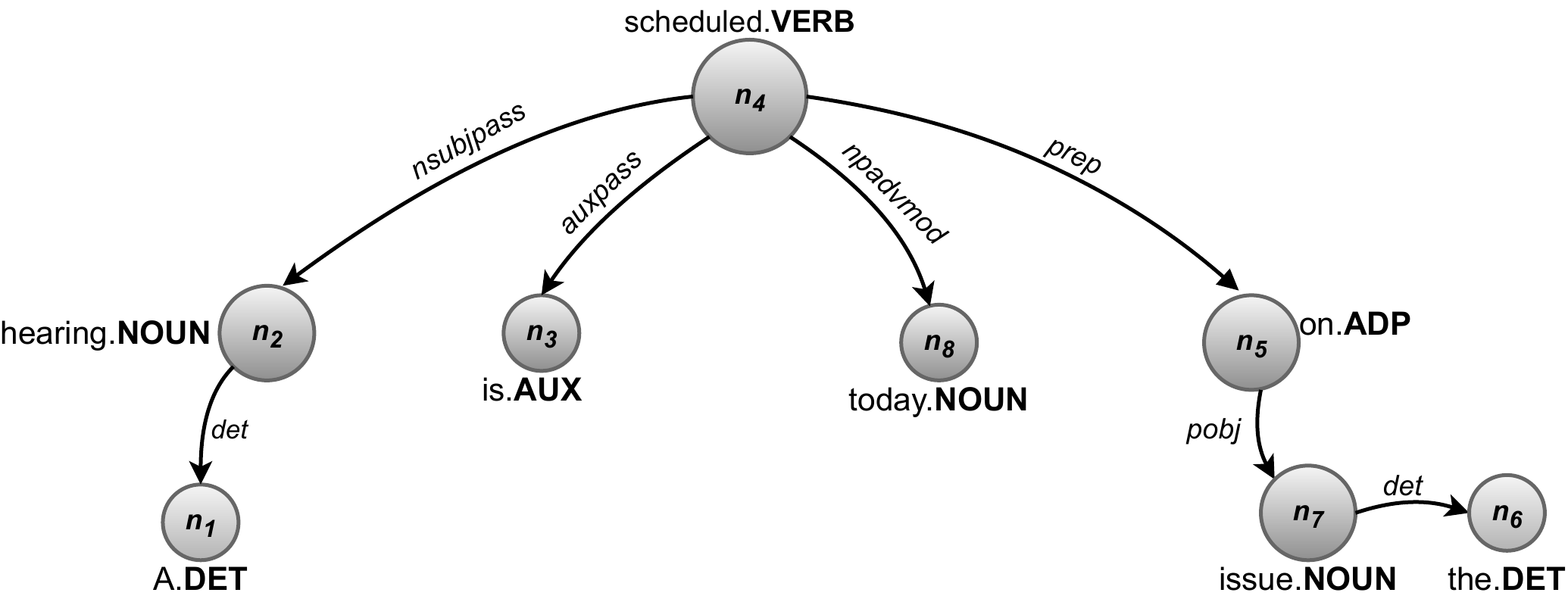}
    \caption{Hierarchy (graph) formulation for DARÉ training: dependency-based tree for the sentence `\textit{A hearing is scheduled on the issue today}'.}
    \label{fig:dare2}
    \vspace{-3mm}
\end{figure}

Dependency parsing is a function that maps a sequence of tokens \{$n_1, n_2, ...$\} to a dependency tree. A dependency parse tree is a directed graph with $N$ nodes and $E$ edges, where each node represents an individual token $n_i$, and each edge represents the syntactic dependency between $n_i$ and $n_j$. An edge $n_i \overset{d_k}{\rightarrow} {n_j}$ comprises a parent node directed at the child node with the dependency $d_k$, where $d_k$ is the nature of the dependency. We employ spaCy\footnote{\url{https://spacy.io/}} for the dependency parsing. An example is shown in Figure \ref{fig:dare2}.

\subsubsection*{Poincaré Embedding:} As highlighted by \citet{nickel2017poincare}, embedding hierarchical graph-based information in Euclidean space can be difficult, owing to the exponential growth of nodes which brings leaf nodes in different branches close to one another thereby distorting hierarchies. With the Poincaré ball, the extent from the center grows exponentially, allowing one to fit an erratic amount of levels in the hyperbolic space. We then optimize our word vectors on this space and optimize the following loss function with negative-sampling:
\begin{equation}
\label{eq:loss}
    L(\theta) = \sum_{(x, y) \in\ \Delta}\log\frac{\exp^{-dist(x,y)}}{\sum_{y' \in\ R(x)}\exp^{-dist(x,y')}}
\end{equation}
where $dist(x,y)$ is the Poincaré distance such that,
\begin{equation}
    dist(x,y) = \text{arccosh}(1 + 2\frac{||x - y||^2}{(1 - ||x||^2)(1 - ||y||^2)})
\end{equation}
where $\theta$ is our set of vectors, $\Delta$ is the set of all embedded hierarchies-cum-dependencies (in this case, $\forall\ n_i \rightarrow n_j$), and $R(x)$ is the set of random tokens that are not associated with $x$. Additionally, we attempt at the disambiguation of part-of-speech (POS) by formulating our $\theta$ as word vectors of the tokens augmented with their POS, such that $x=n_i.\text{POS}_i$. Our loss function $L$ is trained similar to how it would be in Euclidean space. However, the only difference is that we employ Riemann Gradient Descent \cite{bonnabel2013stochastic} for the optimization. We utilize Gensim's implementation\footnote{\url{https://radimrehurek.com/gensim/models/poincare.html}} to train DARÉ.

\subsection{Feature Net (\textit{f-net})}
Prior to discussing \textit{r-net} and \textit{s-net}, we discuss the component \textit{f-net} which is common to both the aforementioned networks and serves as the feature extractor. To put this into context, an \textit{f-net} comprises of stacked BiLSTMs, whose hidden units are then processed by a sequential self-attention mechanism, as suggested by \citet{zheng2018opentag}. 

Inspired by the Inception \cite{Inception:cvpr:2015} architecture, we optimize \textit{f-net} through an auxiliary softmax layer. The intuition behind having these auxiliary outputs is that they would act as implicit assistance against the vanishing gradient problem and make low-level features of the network more accurate. As mentioned before, the transitional outputs (all hidden units) from the BiLSTM are used as inputs for our \textit{r-net} and \textit{s-net}. To emphasize again, each has an individual \textit{f-net} with no shared parameters.

\subsection{Definition Net (\textit{d-net})}
As mentioned earlier, the only distinguishing feature between \textit{f-net}$_s$ and \textit{f-net}$_r$ is that the former comprises of \textit{d-net}.
The \textit{d-net} module helps in aligning the inputs to predefined guidelines/definitions that elucidate the characteristics of claims and non-claims. Intuitively, for any given input text, we find its alignment against the aforementioned sets of definitions using an attention-based mechanism. This helps us draw divergent associations from the input with respect to claims and non-claims.

To put formally, suppose we have two sets of definitions, $C = [c_1, c_2, \cdots, c_n]$ for claims and $NC = [nc_1, nc_2, \cdots, nc_m]$ for non-claims. The input text $t$ forms our query; this query is then processed (discussed below) against each of the definitions in $C$ and $NC$ to get the claim-definition-map and the non-claim-definition-map, respectively. 

The \textit{d-net} comprises of \textit{d-net}$_{C}$ and \textit{d-net}$_{NC}$. These are carbon-copies where one is initialized with the definitions of claims and other with those of non-claims. We process these definitions and input text through a BiLSTM encoder. For each definition, we obtain a \textit{value} vector, and for the input text, we take the last time-step representation as the query vector. Subsequently, for each definition encoding (value$^i$), we calculate its attention score \cite{luong2015effective} against the query. Finally, the query-value-attention-score is pooled (global average) over the sequence axis to obtain a 1-dimensional representation. We repeat the process for each pair of query and definition -- $\langle$\textit{query}, $c_i\rangle \forall i=\{1,2, \cdots, n\}$ and $\langle$\textit{query}, $nc_j\rangle \forall j=\{1,2, \cdots, m\}$. The concatenation of all query-value-attention-scores forms the definition-based representation for $t$ against the respective \textit{definition-set}.

We append the representations from  \textit{d-net}$_C$ and \textit{d-net}$_{NC}$ behind the BiLSTM output at each time step to enhance the feature learning in \textit{f-net}$_{\textit{s}}$ and thereby, explicitly helping the transient BiLSTM features to become more archetypal. The \textit{d-net} is only part of \textit{f-net}$_{\textit{s}}$ owing to the fact that \textit{s-net} is our primary classifier whereas \textit{r-net} acts analogously to a feature-selector. 

\subsection{Regulation Net (\textit{r-net})}
The regulation network acts to collect class-invariant (common, shared amongst the classes) vector representations from \textit{f-net}$_{\textit{r}}$. It functions as a network trained in parallel to the \textit{s-net}. As highlighted in \cite{qin2020feature, ganin2015unsupervised}, we employ a Gradient Reversal Layer (\textit{GRL}) to capture the class-invariant features. In a nutshell, a gradient-reversal layer can be thought of as a pseudo-functional mapping where the forward and backward propagation are respectively defined by two opposed equations as follows:
\begin{equation}
\label{eq:grl}
    \textit{GRL}(x) = x \qquad
    \frac{\delta\textit{GRL}(x)}{\delta x} = -I
\end{equation}

The transient BiLSTM's output from \textit{f-net}$_{\textit{r}}$, that serves as the input to \textit{r-net}, learns the invariant features and is then drifted to \textit{s-net} for computing the orthogonal projection.

\begin{table*}[h]
\centering
\caption{Statistics of four datasets used in our experiments.} 
\vspace{-3mm}
\resizebox{0.8\textwidth}{!}{
\begin{tabular}{l|c|c|c|c|c|c|c|c}
\bf Dataset & \multicolumn{2}{c|}{\bf Twitter} & \multicolumn{2}{c|}{\bf Online Comments (OC)} & \multicolumn{2}{c|}{\bf Web Discourse (WD)} & \multicolumn{2}{c}{\bf Micro Text (MT)} \\ \cline{2-9}
 & \bf Claim & \bf Non-claim &  \bf Claim & \bf Non-claim &  \bf Claim & \bf Non-claim & \bf Claim & \bf Non-claim \\ \hline
 
\hline
\bf Train set & 7354 &	1055 &	623 & 7387 & 190 &	3332 &	100 & 301 \\ 

\bf Test set & 1296 &	189 &	64 & 730 &	14 & 221 &	12 & 36 \\ 
\hline

\bf Overall & 8650 & 1244 & 687 & 8117 & 204 & 3553 & 112 & 337\\
\hline
\end{tabular}}
\label{tab:dataset-stats}
 \vspace{-2mm}
\end{table*}

\subsection{Spotlight Net (\textit{s-net})}

The spotlight network is the prime module of \name. We amalgamate the class-invariant features of \textit{r-net}s through an attention orthogonal project layer ({\em a-OPL}). The orthogonal projection layer aims at drawing the choicest feature representations  \cite{qin2020feature}.

For convenience, we refer to the feature representation from \textit{f-net}$_{\textit{s}}$ and \textit{f-net}$_{\textit{r}}$ as $\textit{v}_s \in \mathbb{R}^{l \times d}$ and $\textit{v}_r \in \mathbb{R}^{l \times d}$, respectively. 

We design \textit{s-net} to extract the semantic representation for our input text $t$ and project the same into a non-homogeneous domain space. To accomplish this, we project $v_s$ onto the orthogonal direction of $v_r$. The space orthogonal to $v_r$ should, in theory, be rid of class-homogeneity, and projecting $v_s$ onto it should lead to discriminative information being stripped of class-invariant knowledge. Further, we describe the mathematical details of \textit{a-OPL}.
The projection between two vectors $u$ and $v$ is defined as,
\begin{equation}
    proj(u,v) = \frac{u.v}{|v|}\frac{v}{|v|}
\end{equation}
Utilizing the above equation, we project $v_s$ onto $v_r$ while tending to each time-step from their respective LSTM using a TimeDistributed Layer (TDL) such that,
\begin{equation*}
\begin{split}
     v_{(s,i),(r,i)} &= \textit{proj}(v_{(s,i)}, v_{(r,i)})\\
     v_{s,r} &= \textit{TDL}_{\forall i}(v_{(s,i), (r,i)})
\end{split}
\end{equation*}

We then find the projection of $v_s$ in the orthogonal direction of $v_{s,r}$ while again tending to each step using a TDL such that,
\begin{equation*}
\begin{split}
    \widetilde{v}_{(s,i)} &= \textit{proj}(v_{(s,i)}, (v_{(s,i)} - v_{(s,i),(r,i)}))\\
    \widetilde{v}_{s} &= \textit{TDL}_{\forall i}(\widetilde{v}_{(s,i)})
\end{split}
\end{equation*}
 
To further refine this feature representation, we then attend to each time-step in $\widetilde{v_{s}}$ using sequential self-attention \cite{zheng2018opentag} such that,
\begin{equation}
\label{eq:attentive}
    attentive_{\widetilde{v}_{s}} =\widetilde{v_{s}}^a = softmax(\sigma(W_a\widetilde{v}_{s,i} + b_a)) 
\end{equation}
The aforementioned forms the basis for our \textit{a-OPL}. The attentive vector $\widetilde{v_{s}}^a$ is then utilized for classification.

We train  \textit{s-net} and   \textit{r-net}   parallel to each other and employ sparse categorical focal loss. In a classification setting, with labels $y$, the loss is defined as,
\begin{equation}
    L(y,\hat{p}) = -(1-\hat{p}_y)^{\gamma}log(\hat{p}_y)
\end{equation}
where $p$ is a vector that represents the approximate probability distribution across our two classes, and $\gamma$ is the \textit{focusing} parameter, which in essence acts to down-weigh easy-to-classify examples. Higher $\gamma$ implies high discounting of the easy-to-classify examples.

\section{Datasets}
Owing to our formulation of a claim detection system for informal texts on the web, we accumulate four publicly-available web-based datasets as mentioned below: 
\begin{itemize}[leftmargin=*]
\setlength{\itemsep}{2pt}
    \item \textbf{Twitter Dataset}: It is a claim detection dataset of COVID-19 tweets released recently \cite{gupta-etal-2021-lesa}. The dataset consists of $\sim{10k}$ labeled tweets. The authors also released a set of definitions for claim and non-claim text in their annotation guidelines\footnote{\url{https://bit.ly/3yAhhW0}}. As mentioned earlier, we utilize them for our definition network (\textit{d-net}). In total, the annotation guidelines consists of 10 claim definitions and 8 non-claim definitions. 
    
    \item \textbf{Online Comments (OC):} \citet{6061427} released a dataset containing comment threads from the blog posts on LiveJournal. 
    
    \item \textbf{Web Discourse (WD):} \citet{habernal-gurevych-2015-exploiting}  released a dataset containing blog posts and user comments stripped to individual sentences and/or phrases.
    
    \item \textbf{Micro Text (MT):} \citet{peldszus-stede-2015-joint} released a corpus of $\sim$500 discussions on controversial issues.
\end{itemize}

Data statistics coupled to the four assimilated datasets are provided in Table \ref{tab:dataset-stats}. 

We can observe that, as opposed to the Twitter dataset, MT, OC and WD datasets are imbalanced towards non-claims. Due to the label skewness, we adopt a sampling mechanism, similar to our precedent work \cite{gupta-etal-2021-lesa}, for our experiments (c.f. Section 5).

\section{Experiments and Results}
In this section, we delineate an exhaustive analysis of our model's performance and also carry out a predictive comparison against the state-of-the-art claim detection systems. We additionally analyze the predictions made by our best model for out-of-sample instances from Twitter. Besides the aforementioned, we conduct an ablation study to evaluate the substance values affixed by each sub-module of our architecture.

\begin{table*}[t]
\centering
\caption{Macro F1 ($m\text{-}F1$) and Claim F1 ($c\text{-}F1$) of the competing models on different web-based datasets. The first two rows determine the effectiveness of DARÉ embeddings over the standard BERT representation. LESA \cite{gupta-etal-2021-lesa} is the current state-of-the-art system for all four datasets.} 
{
\begin{tabular}{l|c:c|c:c|c:c|c:c|c:c}
\bf Model & \multicolumn{2}{c|}{\bf Twitter} & \multicolumn{2}{c|}{\bf Online Comments (OC)} & \multicolumn{2}{c|}{\bf Web Discourse (WD)} & \multicolumn{2}{c|}{\bf Micro Text (MT)} & \multicolumn{2}{c}{\bf Average}\\ \cline{2-11}
 & $m\text{-}F1$ & $c\text{-}F1$ & $m\text{-}F1$ & $c\text{-}F1$& $m\text{-}F1$ & $c\text{-}F1$& $m\text{-}F1$ & $c\text{-}F1$ & $m\text{-}F1$ & $c\text{-}F1$\\ \hline

\hline
K-means -- BERT & 0.46 & 0.70  & 0.39 & 0.09 & 0.41 & 0.04 & 0.49 & 0.35 & 0.44 & 0.30\\ 

K-means -- DARÉ  & 0.52 & 0.83 & 0.39 & 0.16 & 0.47 & 0.11 & 0.50 & 0.21 & 0.47 & 0.33\\ 

\hline
BERT \cite{devlin2018bert} & 0.50 &	0.67 &	0.50 &	0.24 &	0.48 &	0.23 &	0.75 &	0.69 & 0.56 & 0.46\\ 

XLNet \cite{yang2019xlnet} & 0.52 &	0.70 &	0.45 &	0.24 &	0.51 &	0.12 &	0.49 &	0.43 & 0.49 & 0.37 \\ 
\hline

Accenture \cite{williams2020accenture} & 0.48 & 0.15 & 0.44 & 0.16 & 0.34 & 0.11 & 0.48 & 0.28 & 0.44 & 0.18\\

Team Alex \cite{nikolov2020team} & \bf 0.70 & 0.88 & 0.46 & 0.23 & 0.60 & 0.34 & 0.75 & 0.64 & 0.63 & 0.52\\


LESA \cite{gupta-etal-2021-lesa} & 0.67 & 0.89 & 0.51 & 0.26 & 0.61 & 0.35 & 0.80 & 0.71 & 0.65 & 0.55\\ 
\hline

\bf \name & 0.67 &  \bf 0.92  & \bf 0.60 & \bf 0.27 & \bf 0.78 & \bf 0.59 & \bf 0.85 & \bf 0.79 & \bf 0.73 & \bf 0.64 \\ \hline
\end{tabular}}
\label{tab:results}
\end{table*}

\subsection{Experimental Setup}
In this section, we layout the backdrop for our experiments and highlight the key conditions and practices.

To compute the DARÉ embedding of size $100$, we use the open source Sentiment140 corpus, comprising 1.6 million tweets\footnote{\url{https://www.kaggle.com/kazanova/sentiment140}}. 
We use stacked BiLSTMs with 256 hidden units for both \textit{f-nets}. To emphasize again, there are no shared parameters between \textit{f-net}$_s$ and \textit{f-net}$_r$. Additionally, to encode our query and definitions in the \textit{d-net}, we use a BiLSTM with 64 hidden units. We use pre-furnished definitions proposed in our previous work \cite{gupta-etal-2021-lesa} to harbour our \textit{d-net}; we encipher \textit{d-net}$_C$ with $10$ definitions, and \textit{d-net}$_{NC}$ with $8$ definitions.

To train our model, we proceed with a vocabulary size of $30k$, and a maximum document length of 50. We use the Adam \cite{kingma2014adam} optimizer and the sparse categorical focal loss \cite{lin2017focal}. We train the model for $100$ epochs with a batch size of $32$ and exercise early stopping. Since most of the datasets are unbalanced towards one class, we adopt the sampling technique to alleviate the issue. We experiment with multiple sampling ratio (c.f. Table \ref{tab:ablation}), and as a result of fine-tuning, we select a sampling ratio of 5:2 (claims: non-claims) for the Twitter dataset based on foraging through the sampling ratio search space. For consistency, we select a sampling ratio of 5:2 (non-claims: claims) for the OC, WD and MT datasets as well. Note that the sampling technique is in line with the previous neural attempts that procured best results on a 1:1 ratio \cite{daxenberger2017essence, gupta-etal-2021-lesa}. Additionally, to ensure the riddance of seed stochasticity and to incorporate maximal data, we train for 5 random splits and average them using voting. Also, to incorporate a more holistic weighting mechanism, we vote on predictions across three different values (1, 2, 3) of the $\gamma$ parameter in the focal loss. For evaluating the claim detection systems, we compute {\bf claim-F1 ($c\text{-}F1$)} and {\bf macro-F1 ($c\text{-}F1$)} scores.

Another fact of importance is that with \name, we design an architecture which is lighter in comparison to its previous state-of-the-art systems, especially LESA. LESA churns approximately $111M$ model parameters, while our proposed model has only $7M$ model parameters. Moreover, the standard XLNet and BERT-based models require a few $100M$ parameters for the same. 

As is evident, constructing models in adherence with the task statement at hand can be equally as effective if not more, and at times can be accomplished with a fraction of the compute.

\subsection{Baseline Models}
Due to highly subjective nature of claims, claim detection can more often than not prove to be a demanding task even for humans let alone machines. As with automated, neural (machine-based) claim detection, the problem becomes even more acute in case of web-based short texts, that usually lack soundness in their linguistic edifice. Most of the existing claim detection models (including state-of-the-art systems) struggle with the accurate identification of claims. To assess and contrast the performance of  \name, we consider the following systems as baselines: 
\begin{itemize}[leftmargin=*]
\setlength{\itemsep}{2pt}
    \item \textbf{BERT} \cite{devlin2018bert}: It is a bidirectional transformer-inspired auto-encoder LM that we fine-tune for classification.
    \item \textbf{XLNet} \cite{yang2019xlnet}: Similar to BERT, this too is a bidirectional transformer-inspired LM, the only difference being that this is an auto-regressive LM. We fine-tune it for classification.
    \item \textbf{Accenture} \cite{williams2020accenture}: The authors employed a fine-tuned RoBERTa-based system and nabbed the first position in the CLEF-2020 claim detection shared task \cite{checkthat:clef:2020}.
    \item \textbf{Team Alex} \cite{nikolov2020team}: The system ranked second at CLEF-2020 shared task. The authors proposed the fusion of RoBERTa-based features and Twitter meta-data to detect claims.
    \item \textbf{LESA} \cite{gupta-etal-2021-lesa}: This is the state-of-the-art claim detection system wherein the authors proposed a system that leverages part-of-speech and dependency-based linguistic encoders  in sync with a BERT-based encoder to detect claims.
    \item In addition, we also perform a simple \textbf{K-means} clustering-based evaluation. We assign dataset points to one of the two clusters -- claim and non-claim, considering their BERT and Poincaré representations, separately. 
\end{itemize}

\subsection{Performance Comparison}
We present our collated results in Table \ref{tab:results}. To compute the base efficacy of the DARÉ embedding, we compare it against the BERT embeddings \cite{reimers-2019-sentence-bert} without proffering any external supervision. Please note that the Sentence-Transformers package\cite{reimers-2019-sentence-bert} facilitates out-of-box computation of BERT-based dense vector representations for sentences\footnote{https://www.sbert.net/}. We employ K-means clustering to segregate the claim with non-claim clusters. We evaluate the test data points in both clusters and report the results in the first two rows of Table \ref{tab:results}. We observe that DARÉ performs better than BERT on the Twitter dataset by a considerable margin. We also perform K-means clustering on the remaining three datasets as well and observe improvements in most of the cases. The improvements could possibly be attributed to our trained distribution being closer to web-based informal texts despite the training corpus being significantly smaller than BERT's (Wikipedia: $2,500$ million words, Book Corpus: $800$ million words). 

Furthermore, we evidently observe that \name\ outperforms all the existing baseline systems including the current state-of-the-art, LESA \cite{gupta-etal-2021-lesa}. On the Twitter dataset, \name\ obtains the foremost $c\text{-}F1$ score in contrast to all other baseline systems -- it accounts for a $+3.3\%$ improvement over LESA in $c\text{-}F1$. With the OC dataset, we find that all the baseline systems including \name\ report low scores for claims. However, \name\ does yield $m\text{-}F1$ of 0.60 (with $+9$ points improvement over LESA's performance) -- suggesting it performs well for the non-claim class. Out of the four datasets, we observe the highest relative improvement on the WD dataset with   $0.59$ $c\text{-}F1$ and  $0.78$ $m\text{-}F1$, translating to a climb of $68.5\%$ and $27.8\%$ over LESA, respectively. On the MT dataset as well, we observe an increment of $11.26\%$ in $c\text{-}F1$ and $6.25\%$ in $m\text{-}F1$.

On average, \name\ improves the state-of-the-art performance by $16.36\%$ in $c\text{-}F1$ and by $13.3\%$ in $m\text{-}F1$. As a general observation, we see how systems grounded in linguistics, such as \name\ and LESA outperform large LMs like BERT and XLNet, which in turn goes to indicate the importance of task-specificity in model adaptation.

\begin{table}[!t]
\centering
\caption{Ablation result for \name\ on the Twitter dataset. The symbol ($-$) signifies the absence of respective module. Claim-F1 ($c\text{-}F1$), Non Claim-F1 ($nc\text{-}F1$), Macro-F1 ($m\text{-}F1$) and Weighted-F1 ($w\text{-}F1$) are reported. Sampling ratio shows our attempt to alleviate the label skewness.} 
\resizebox{\columnwidth}{!}
{
\begin{tabular}{l|c|c|c|c|c}

    & \textbf{Sampling ratio} & & & &\\ 
    \bf Ablation & [Claim:Non-Claim]  & \bf $c\text{-}F1$ & \bf $nc\text{-}F1$ &	\bf $m\text{-}F1$ & \bf $w\text{-}F1$ \\ 
    \hline
    \multirow{3}{*}{\name} & Org & \bf 0.92 & 0.38 & 0.65 & 0.85 \\ \cdashline{2-6}
    & $[1:1]$ & 0.89 & 0.33 & 0.61 & 0.81 \\ \cdashline{2-6}
    & $[5:2]$ & \bf 0.92 & \bf 0.41 & \bf 0.67 & \bf0.86 \\
    \hline
    \multirow{2}{*}{$\quad-$ \{\textit{focal loss, d-net}\}} & $[1:1]$ 	& 0.85  & 0.36 &	0.60 &	0.78 \\ \cdashline{2-6}
    & $[5:2]$  &	0.90 & 0.39	& 0.65 &	0.83 \\
    \hline
    \multirow{2}{*}{$\quad-$ \{\textit{focal loss}\}} & $[1:1]$  &	0.86 & 0.37 &	0.61 &	0.80 \\ \cdashline{2-6}
    & $[5:2]$ & 0.91 & \bf 0.41 & 0.66 &	0.84 \\ \hline
    \multirow{2}{*}{$\quad-$ \{\textit{DARÉ}\} $+$ \{\textit{GloVe}\}} & $[1:1]$ &	0.90 & 0.38 & 0.64 & 0.83 \\ \cdashline{2-6} 
     & $[5:2]$ & 0.91 & 0.36 & 0.63 & 0.84 \\ \hline 
\end{tabular}}

\label{tab:ablation}
\vspace{-2mm}
\end{table}

\begin{table*}[h]
\centering
\caption{Error analysis of the outputs on various datasets. For comparison, we also show the predictions of the current state-of-the-art system, LESA \cite{gupta-etal-2021-lesa}.}
\label{tab:error-analysis}
\vspace{-2mm}
{
\begin{tabular}{c|c|p{33em}|c|c|c}

  \multicolumn{2}{c|}{\multirow{2}{*}{\bf Dataset}} & \multirow{2}{*}{\bf Example} & \multirow{2}{*}{\bf Gold} & \multicolumn{2}{c}{\bf Prediction} \\ \cline{5-6}
   \multicolumn{2}{c|}{} & & & \bf \name & \bf LESA \\ \hline

\hline

    
         
    & $t_{1}$ & \textit{RT @PirateAtLaw: No no no. Corona beer is the cure not the disease https://t.co/fnba2fr2m2} & claim & claim & \textcolor{red}{non-claim} \\ \cdashline{2-6}

    \multirow{2}{*}{TWR} & $t_{2}$ & \textit{@zlj517 Vaccine development is urgent. Please let me know when an effective vaccine is completed in the world. I go get it and bring it to China. Wait for early development. \#pichperfact} & non-claim & \textcolor{red}{claim} & \textcolor{red}{claim}\\ \cdashline{2-6} 
     
    
    & $t_{3}$ & \textit{\#China has just discovered \#vaccine against \#coronavirus. Thank you china} & claim & claim & \textcolor{red}{non-claim}\\ \hline
    
    
    & $t_{4}$ & \textit{I usually walk around and attempt to read all the names on the old gravestones.} & non-claim & \textcolor{red}{claim} &  non-claim \\ \cdashline{2-6}
    
    \multirow{2}{*}{OC} & $t_{5}$ & \textit{Casey Siemezko is Charlie in Young Guns...which bothers me because I keep wantin to see him without his beard/mustache and this is the closest thing to it, but he has his 3D glasses...cant win.} & non-claim & \textcolor{red}{claim} &  \textcolor{red}{claim}\\ \cdashline{2-6} 
    
    & $t_{6}$ & \textit{My only big smile moment was Kurt's dad knowing Kurt was g*y since he was a t*t, lovely little scene, playing against the dad's macho image.} & claim & claim & \textcolor{red}{non-claim}\\ \hline
    
    & $t_{7}$ & \textit{I don't know about the rest of Virginia, but in Northern Virginia we have excellent public schools.} & claim & \textcolor{red}{non-claim} & \textcolor{red}{non-claim} \\ \cdashline{2-6}

    \multirow{2}{*}{WD} & $t_{8}$ & \textit{I am a 12-year-old 6th grader turning 13 in May, so I know what it’s like.} & non-claim & non-claim & \textcolor{red}{claim}  \\ \cdashline{2-6}
    
    & $t_{9}$ & \textit{Here is my experience and I hope it will help you with your decision: In preschool, I had some issues, just like your son.} & non-claim & \textcolor{red}{claim} & \textcolor{red}{claim}\\ \hline 
    
    
    & $t_{10}$ & \textit{That's why Germany should not introduce capital punishment!} & claim & \textcolor{red}{non-claim} & claim \\ \cdashline{2-6}

    \multirow{2}{*}{MT} & $t_{11}$ & \textit{Alternative treatments should be subsidized in the same way as conventional treatments, since both methods can lead to the prevention, mitigation or cure of an illness.} & claim & claim & \textcolor{red}{non-claim} \\ \cdashline{2-6}
    
    & $t_{12}$ & \textit{Besides it should be in the interest of the health insurers to recognize alternative medicine as treatment, since there is a chance of recovery.} & non-claim & \textcolor{red}{claim} & \textcolor{red}{claim}\\ \hline \multicolumn{6}{c}{}\\
    
\end{tabular}}

\vspace{-3mm}
\end{table*}

\subsection{Ablation Study}
Given that we hinge our model around Twitter, we perform the ablation study by hacking off individual components from \name\ one at a time, and thereafter, we evaluate the same on the Twitter dataset \cite{gupta-etal-2021-lesa}. We present the ablation study results in Table \ref{tab:ablation}. We report  $c\text{-}F1$ along with   $m\text{-}F1$ and weighted-F1 ($w\text{-}F1$).

We draw the simplest variant of \name\ by dropping \textit{d-net} and employing sparse categorical cross-entropy in place of focal loss to train our model. Additionally, along with the mentioned withdrawals, we train on a 1:1 sampling in place of the 5:2 sampling. As can be seen in Table \ref{tab:ablation} (rows 4-5), we observe an increase of 5 $c\text{-}F1$ points and 5 $m\text{-}F1$ points simply by reverting back to the 5:2 sampling. We conduct another similar ablation, while retaining \textit{d-net} with categorical cross-entropy. We again observe an increase of 5 $c\text{-}F1$ points and 5 $m\text{-}F1$ points on a reversion to the 5:2 sampling. It is worth mentioning that the previous benchmark on the Twitter dataset applied a 1:1 sampling across all their experiments \cite{gupta-etal-2021-lesa}. We, however, espy worse results on using the same. We additionally discern that in two distinct cases, the addition of  \textit{d-net} results in a boost -- we see a boost of 1  $c\text{-}F1$ point and 1 $m\text{-}F1$ point when comparing ablations that differ in their residence of   \textit{d-net}.

Within \name\, we experiment with different sampling ratios, as is clearly evident we outperform on the original sampling and the 1:1 sampling. We observe that with the original sampling, we see competitive results on $c\text{-}F1$, however, the same doesn't hold true for $nc\text{-}F1$, thereby the 5:2 sampling is a better fit, given that in addition to better results, it also helps keep the skew subservient. 

Furthermore, we detect that the variant of \name\ that comes without focal loss, performs worse in comparison to \name; the $c\text{-}F1$ and the $m\text{-}F1$ values drop by 1 point each. Another interesting ablation is where we choose to initialize \name\ with GloVe \cite{pennington2014glove} in place of DARÉ. We see that \name\ in its native state outperforms GloVe initialization by 1 $c\text{-}F1$ point and 4 $m\text{-}F1$ points (rows 8-9).

\subsection{Error Analysis} 
\label{sec:comp-err-analysis}
Glancing at Table \ref{tab:results}, one can evidently infer that all the claim detection systems including \name\ are still far from absolute and are therefore prone to errors. To reinforce our point once again, owing to the highly impressionistic nature of claims and folksiness on OSM sites, the detection of claims is cumbersome. To qualitatively appraise the performance of   \name, we perform error analysis in this section. Table \ref{tab:error-analysis} highlights a few randomly sampled instances from the Twitter, OC, WD, and MT datasets, along with their gold and output labels as predicted by \name. For comparison, we analyze the predictions from the state-of-the-art claim detection architecture, LESA \cite{gupta-etal-2021-lesa} as well. In some cases, both \name\ and LESA fail to identify claims; however, in most of the cases \name\ performed well. We additionally report instances misclassified by \name\ and/or LESA. 

\begin{table*}[t]
\centering
\caption{Human evaluation on random tweets (non-dataset examples). \name\ reports macro-F1 of 0.60 and claim-F1 of 0.73 on 50 random samples.}
\label{tab:predictions}
{
\scalebox{0.99}{
\begin{tabular}{c|p{42em}|c|c}

    & \bf Example & \bf \name & \bf Human \\
\hline

\hline
    $t_{1}$ & \textit{\#CovidVaccine | Assam vaccinated only 15\% residents, young struggle to book slots. NDTV's Ratnadip Choudhury reports} & claim & claim \\ \hline
    
    
    $t_{2}$ & \textit{We went from May 1 to May 17 in 2 day} & \textcolor{red}{claim} & non-claim\\ \hline 

    $t_{3}$ & \textit{Nearly 51 lakh COVID-19 vaccine doses will be received by states/UTs within next 3 days: Health ministry.} & claim & claim\\ \hline 
     
    $t_{4}$ & \textit{I STAND WITH HUMANITY \
    \#IndiaStandwithPalestine} & non-claim & non-claim\\ \hline 
      
    
    $t_{5}$ & \textit{Par for the course. As if we'd trust an internal review. We are asking for \#COVIDPublicEnquiryNow} & \textcolor{red}{non-claim} & claim\\ \hline 
       
    $t_{6}$ & \textit{They claim they are no longer asking for Aadhaar/mobile to give food at Indira canteens. But our Youth Congress team found otherwise. And documented it. We will continue to expose this government’s lies.} & claim & claim\\ \hline 
    
    $t_{7}$ & \textit{A third Australian has died in India from COVID-19. The family members of 11,000 stranded Aussies are pleading for the government to bring them home. \#9News} & claim & claim\\ \hline
    
    $t_{8}$ & \textit{My heartfelt gratitude to the men in uniform who did not deter from putting their lives in danger saving the lives of our citizens under extreme conditions.} & non-claim & non-claim\\ \hline
    
    $t_{9}$ & \textit{Interacted with doctors across India. They shared insightful inputs, based on their own experiences of curing COVID-19. The determination of our doctors during these times is remarkable!} & \textcolor{red}{claim} & non-claim\\ \hline
    
    $t_{10}$ & \textit{Abolish unpaid internships. There is absolutely no valid reason that justifies why you’re having students work 40h/week and paying them nothing.} & claim & claim\\ \hline
\end{tabular}}}
\end{table*}

As underlined previously, on the Twitter dataset, we observe that \name\ obtains better classification results in comparison to all other baseline systems. In examples $t_1$ and $t_3$, we see that LESA misclassifies both the examples. In comparison, \name\ identifies the assertion and classifies them correctly. With mistakes being inevitable, we observe that most of the misclassified samples are non-claim ($t_2, t_4, t_5, t_9, t_{12}$). A potential reason could be the skewed nature of the Twitter dataset \cite{gupta-etal-2021-lesa} -- the dataset is imbalanced with a hulking predisposition towards claims (the number of claims is greater even after the 5:2 sampling ratio). The presence of the phrase, \textit{`effective vaccination is completed'} in example $t_{2}$, drives the system to assert on it and incorrectly predicting it as a claim. Example $t_7$ has a dearth of context and could possibly have been a part of a bigger phrase, and on top of that it is not a statement that would severely affect public opinion, which is presumably why \name\ misclassifies it. The latter argument was based off the guidelines proposed in our previous work \cite{gupta-etal-2021-lesa}. Clearly, the gold label for $t_9$ indicates that it is not a claim; however, within the realm of possibility, it does seem that the person indirectly claims about having issues. It is possible that \name\ misclassifies in this case owing to this dormant pattern. Sentence $t_{12}$ emphasizes a chance of recovery with an alternative medicine, i.e., they report a medical fact to be true, albeit with chance. There exists a possibility that \name\ and LESA interpret a chance of recovery using medication as a claim, and therefore, misclassify it.

\subsection{Human Evaluation in the Wild}  
Over time, OSM sites have emerged as the hub for short, unstructured pieces of informal text, where the amount of slang and incoherence in writing is generally more significant than other online platforms. Considering the prime focus of \name, we tend to evaluate our \name's performance against real-world data to detect claims on the web. We collect 50 random samples from Twitter and predict their labels (claim or non-claim) using \name. Note that we collect these examples in the wild. Subsequently, we present the predictions to three human evaluators\footnote{They are  linguistics by profession; and their age ranges between $24$-$45$ years.} and ask them to verify the labels following the claim annotation guidelines defined in \cite{gupta-etal-2021-lesa}. Finally, a majority voting was used to get the final gold-label for these 50 tweets. We obtain an inter-annotator score (\textit{Fleiss kappa}) of 0.76 among the three evaluators.

We present some of the instances in Table \ref{tab:predictions}. As expected, most of the claims were correctly labelled by \name. Out of 50 samples, \name\ classifies 32 samples correctly. We observe $m\text{-}F1$ score of 0.60, while $c\text{-}F1$ score of 0.73. Along with that, our model marked some false positives at the expense of its precision. This is not an ideal situation but a better scenario than biasing towards false negatives where claims are wrongly classified as non-claims. 

Examples $t_{1}$ and $t_{3}$ are claims exhibiting some statistics. Our model \name\ rightly captured these values and classified both the tweets as claims. However, \name\ was unable to learn the importance of numerical features in $t_{2}$, specially when they occurred several times within the text, where it failed to interpret the significance of the numerical features and ended up mislabelling the tweet.  
Following the claim guidelines, we published in our prior work \cite{gupta-etal-2021-lesa}, negating a false claim also accounts as a claim. In example $t_{6}$, the user tries to negate a claim and further claims to document the evidence. \name\ comprehends the assertions and rightly labels as claim. Through example $t_{8}$, the user imparts his/her personal beliefs that might or might not affect the public; thus, \name\ possibly interpreted it as a personal opinion and marked it as non-claim. Examples $t_{7}$ and $t_{10}$ encompass strong claim phrases \textit{`has died in India'} and \textit{`paying them nothing'}, respectively. Clearly, these two examples make strong social assertions that would be of interest to a larger audience and is possibly why it is labeled as claim. Finally, in example $t_9$, the user commends the determination of doctors during the global pandemic situation and expressed their experiences. This example would not fall under the claim category; however, \name\ mis-classifies it as a non-claim possibly due to the presence of the phrase `\textit{curing COVID-19}'.  

Our observations from the in-the-wild evaluation suggest that that   \name\ can assign labels to unseen tweets quite efficiently and accurately. Moreover, we do not follow any unexpected behavior of \name. Thus, furnishing us with empirical shreds of evidence that \name\ can be used for claim detection in informal texts.

\section{Conclusion}
Through this far-reaching and orderly study, we tend to make notable contributions that will end up being considerable strides in the field of claim detection.  
We epitomized Poincaré embeddings with the NLP task, which showed promising results for the claim detection task. The proposed model, \name, determined the existence of claims in the online text by aligning the query to encoded definitions and projecting them into a purer space.   
We evaluated \name\ across four web-based datasets, which comprise short informal texts, and observed up to par results. The comparative investigation espoused the more nuanced performance of our model compared against various existing systems. Experiments demonstrated the superiority of our model with $\ge$3\% $claim\text{-}F1$ improvements over the existing state-of-the-art claim detection system, LESA.
Additionally, we exhibited every individual component's performance and significance in our model through an exhaustive ablation study. Finally, we showed the robustness of \name\ through a qualitative human evaluation in the wild on 50 random samples. 

\section{Acknowledgement}
The work was partially supported by the Accenture Research Grant. T. Chakraborty would like to acknowledge the support of Ramanujan Fellowship, CAI, IIIT-Delhi and ihub-Anubhuti-iiitd Foundation set up under the NM-ICPS scheme of the Department of Science and Technology, India.
\bibliographystyle{ACM-Reference-Format}
\bibliography{bibliography}

\end{document}